\documentclass[nohyperref]{article}

\usepackage{microtype}
\usepackage{graphicx}
\usepackage{subfigure}
\usepackage{booktabs} %

\usepackage{hyperref}

\usepackage[accepted]{icml2022}

\usepackage{amsmath}
\usepackage{amssymb}
\usepackage{mathtools}
\usepackage{amsthm}

\usepackage[capitalize,noabbrev]{cleveref}

\theoremstyle{plain}
\newtheorem{theorem}{Theorem}[section]

\theoremstyle{definition}
\newtheorem{definition}[theorem]{Definition}

\theoremstyle{remark}

\usepackage{xspace}
\usepackage{amssymb}
\usepackage{amsmath}
\usepackage{bm}

\usepackage{color}
\usepackage{algorithm}
\usepackage{algorithmicx}
\usepackage[noend]{algpseudocode}

\newcommand{\expt}{\mathop{\mathbb{E}}}
\newcommand{\bR}{\mathbb{R}}

\newcommand{\cF}{\mathcal{F}}
\newcommand{\cS}{\mathcal{S}}
\newcommand{\cA}{\mathcal{A}}
\newcommand{\cO}{\mathcal{O}}
\newcommand{\cP}{\mathcal{P}}
\newcommand{\cR}{\mathcal{R}}
\newcommand{\cX}{\mathcal{X}}
\newcommand{\cU}{\mathcal{U}}

\newcommand{\cD}{\mathcal{D}}

\newcommand{\fD}{\mathfrak{D}}
\newcommand{\fB}{\mathfrak{B}}
\newcommand{\np}{\Pi_{\theta, \Sigma}}

\newcommand{\fpsron}{\cF_{\textsc{PSRO-N}}}
\newcommand{\neupl}{NeuPL\xspace}
\newcommand{\nash}{Nash\xspace}
\newcommand{\rps}{{\it rock-paper-scissors}\xspace}
\newcommand{\rws}{{\it running-with-scissors}\xspace}
\newcommand{\psro}{{\sc PSRO}\xspace}
\newcommand{\psronash}{{\sc PSRO-Nash}\xspace}

\newcommand{\eqdef}{\mathrel{\mathop:}=}
\newcommand\Tau{\mathcal{T}}

\newcommand{\abr}{\textsc{ABR}\xspace}

\usepackage[textsize=tiny]{todonotes}

\icmltitlerunning{Simplex NeuPL: Any-Mixture Bayes-Optimality in Symmetric Zero-sum Games}

\begin{document}

\twocolumn[
\icmltitle{Simplex Neural Population Learning: \\ Any-Mixture Bayes-Optimality in Symmetric Zero-sum Games}

\begin{icmlauthorlist}
\icmlauthor{Siqi Liu}{ucl,deepmind}
\icmlauthor{Marc Lanctot}{deepmind}
\icmlauthor{Luke Marris}{ucl,deepmind}
\icmlauthor{Nicolas Heess}{deepmind}
\end{icmlauthorlist}

\icmlaffiliation{ucl}{University College London, UK}
\icmlaffiliation{deepmind}{DeepMind, UK}

\icmlcorrespondingauthor{Siqi Liu}{liusiqi@google.com}

\icmlkeywords{Multiagent, Game Theory, Reinforcement Learning}

\vskip 0.3in
]

\printAffiliationsAndNotice{}  %

\begin{abstract}
Learning to play optimally against any mixture over a diverse set of strategies is of important practical interests in competitive games. In this paper, we propose simplex-\neupl that satisfies two desiderata {\em simultaneously}: i) learning a population of strategically diverse basis policies, represented by a single conditional network; ii) using the same network, learn best-responses to {\em any} mixture over the simplex of basis policies. We show that the resulting conditional policies incorporate prior information about their opponents effectively, enabling near optimal returns against arbitrary mixture policies in a game with tractable best-responses. We verify that such policies behave Bayes-optimally under uncertainty and offer insights in using this flexibility at test time. Finally, we offer evidence that learning best-responses to any mixture policies is an effective auxiliary task for strategic exploration, which, by itself, can lead to more performant populations.
\end{abstract}

\section{Introduction}
\label{introduction}

How could we train agents to perform optimally against arbitrary mixtures over diverse opponent policies?
Population learning offers one potential answer: generate a diverse set of opponents and train the agent to respond to mixtures of opponents over the population. 
The question then becomes how the population is generated and what properties it should have.
In two-player zero-sum games, there is a well-known solution to this problem based on game-theoretic foundations: a Nash equilibrium distribution (NE, \citet{ne_jnash}) over a population of policies maximizes an agent's worst-case return against all possible opponent policies. 
Despite its theoretical appeal, searching the entire policy space quickly becomes intractable for most games. To this end, empirical game-theoretic analysis (EGTA, \citet{wellman2006methods}) proposed to study strategic exploration in games by investigating empirical (meta-)games, where each player considers only a small subset of possible policies. Policy-Space Response Oracles (PSRO, \citet{lanctot2017unified}), further proposed a general, iterative framework towards constructing such empirical games. At each iteration, the policy population incorporates a new basis policy that is trained to best-respond to a mixture over its predecessors, following a meta-strategy solver (MSS). Importantly, when the best-response operator is exact, certain meta-strategy solvers produce meta-strategies known to converge to an NE of the game.

One property of the NE target distribution is that it optimizes a safe objective: it maximizes the expected payoff in the {\em worst-case}, with the assumption that the opponents would play minimax-optimally. This assumption, however, rarely holds in practice --- real-world agents could play arbitrarily far from NE, a phenomenon frequently observed among human players \citep{Wright2017Predicting}, due to inadequate training or simply, to the overwhelming complexity of the game. This translates to the unfortunate situation where NE, though unexploitable, often leads to sub-optimal decision making at test time. 
The flexibility for players to express subjective beliefs over the opponent and to play optimally, based on such beliefs is thus of interests. We refer to this ability to play optimally against any mixture over a diverse set of policies as {\em any-mixture optimality}.
Indeed, skilled human players are observed to resort to such flexibility when competing in games, adjusting their behaviours based on assumptions about their opponents so as to play optimally, if their assumptions prove correct \citep{doi:10.1126/science.1108062, schlicht2010human}.

Unfortunately, existing population learning algorithms such as \psro precisely lack such flexibility. The choice of MSS not only controls the strategic diversity of the resulting population, but also, restricts the set of basis policies that can be executed at test time. In particular, the output of population learning is a set of best-responses to specific mixture policies, enumerated by the MSS at each iteration.
Consequently, a player can only play optimally against a few sets of opponents, or forgo optimality entirely and execute the NE mixture policy so as to be assured of safety, in expectation, over many games. At its extreme, a player cannot guarantee to play optimally even when the opponent uses the same population of policies and publicly declares their strategy in advance, nor can they play optimally if they wish to consider all strategies equally likely {\em a priori} without unduly ruling out any opponent strategy.

Our goal is therefore to extend game-theoretic population learning algorithms so as to offer {\em any-mixture optimality} at test time. To this end, we interpret \psro geometrically as iteratively expanding a population simplex whose vertices correspond to the set of basis policies, each best-responding to a point within the simplex from the previous iteration (Figure~\ref{fig:population_simplex}). To instead learn best-responses to {\em all} points within the population simplex, we further generalise recent work on Neural Population Learning (\neupl, \citet{liu2022neupl}), a general framework that incorporates principled population learning algorithms, using scalable and efficient representation for the population of policies via a single conditional neural network. The result is thus a simple extension that not only retains the efficiency and game-theoretic properties of \neupl, but also yields a conditional policy that behaves optimally against {\em arbitrary} mixtures over the policy population {\bf (Section~\ref{sec:methods})}. Additionally, we recognize best-response solving across the population simplex as optimising a continuum of Bayes-optimal objectives \citep{humplik2019meta, ortega_meta-learning_2019} and demonstrate properties of the resulting policies typically associated with Bayes-optimality. In particular, we show that the resulting conditional policies effectively incorporate prior information about their opponents so as to achieve near optimal returns against arbitrary mixtures policies in a game with tractable best-response solutions {\bf (Section~\ref{sec:goofspiel})}. We further compare different choices of policies at test time in a more complex, partially-observed, spatiotemporal strategy game and show that executing the NE mixture policy can be far from optimal whereas executing an {\em uninformed} policy that considers all opponent strategies equally likely {\em a priori} can be highly effective {\bf (Section~\ref{sec:rws})}. Lastly, we show that simplex-\neupl is not only critical in providing {\em any-mixture optimality}, but also, facilitates strategic exploration by promoting transfer across best-responses to the continuum of mixture policies, leading to more performant populations at no extra costs {\bf (Section~\ref{sec:ablation})}.

\section{Background}
\label{sec:background}

\subsection{Partially-Observed Stochastic Games (POSG)}

Stochastic games~\citep{shapley_stochastic_1953} generalise the basic formalism of Markov Decision Processes (MDPs) to multiple players. To model partial observability, we define a symmetric zero-sum partially-observed stochastic game~\cite{Hansen04} by $(\cS, \cO, \cX, \cA, \cP, \cR)$ where $\cS$ defines the state space, $\cO$ the observation space and $\cX: \cS \rightarrow \cO \times \cO$ the observation function that returns partial views of the state for both players. Let $\cP: \cS \times \cA \times \cA \rightarrow \Pr(\cS)$ be the state transition distribution given a state and joint actions, $\cR: \cS \rightarrow \bR \times \bR$ the reward function defining rewards for both players in state $s_t$, denoted $\cR(s_t) = (r_t, -r_t)$. In state $s_t$, players act according to policies conditioned on their respective observation histories $(\pi(\cdot|o_{\le t}), \pi'(\cdot|o'_{\le t}))$. In practice, observation history can be represented as fixed-size embedding with the use of a learned recurrent neural network. Player $\pi$ achieves an expected return of $J(\pi, \pi') = \expt_{\pi, \pi'}[\sum_t r_t]$ against $\pi'$. A game is said to be {\em symmetric} if the expected return of a policy is only dependent on the policy played by the other player, rather than the identity or order of the player. A policy $\pi^{*}$ is said to best-respond to $\pi'$ if $\forall \pi, J(\pi^{*}, \pi') \ge J(\pi, \pi')$. We note $\pi^* \gets \textsc{BR}(\pi')$, if a best-response policy against $\pi'$ can be computed tractably. In practice, an exact best-response operator may be intractable computationally and we define approximate best-response (\abr) operators as $\hat{\pi} \gets \abr(\pi, \pi')$ such that $J(\hat{\pi}, \pi') \ge J(\pi, \pi')$. In other words, an approximate best-response operator produces a policy $\hat{\pi}$ that performs at least as well as $\pi$ against $\pi'$. It's worth noting that we focus on POSG instead of Normal-form Games (NFGs) as our focus is on developing Bayes-optimal policies that can benefit from information gathered through sequential interactions.

\subsection{Population Learning}
Population Learning defines an iterative procedure for strategic exploration in games. In particular, we consider the formalism of Policy-Space Response Oracles (\psro, \citet{lanctot2017unified}) which combined EGTA with deep reinforcement learning. Given a symmetric zero-sum, partially-observed, stochastic game where each player has access to the same set of $N$ policies $\Pi \eqdef \{\pi_i\}^{N-1}_{i=0}$, we define a normal-form empirical (meta-)game where players' $i$-th action corresponds to executing policy $\pi_i$ for an episode. A probability assignment $\sigma \in \Delta^{N-1}$ over the policy population therefore defines a meta-game mixture strategy, or a mixture policy $\Pi^\sigma$ in the underlying game, with $\Delta^{|\Pi| - 1}$ representing the space of $|\Pi|$-dimensional distributions, or the volume of a $(|\Pi| - 1)$-simplex. When executing a meta-game mixture strategy, an action of the meta-game, or a policy in the underlying game, is sampled at the start of each episode, following $\sigma$. The definition of (approximate) best-response readily extends to mixture policies, with $J(\pi, \Pi^\sigma) = \expt_{i \sim \sigma}\Big[\expt_{\pi, \pi_i}[\sum_t r_t]\Big]$. We further define the empirical payoff matrix $\cU \in \bR^{|\Pi| \times |\Pi|} \gets \textsc{Eval}(\Pi)$ with $\cU_{ij} \eqdef J(\pi_i, \pi_j)$ the payoff of the $i$-th meta-game pure-strategy when playing against the $j$-th. We further recall the definition of meta-strategy solver (MSS) $f: \bR^{|\Pi| \times |\Pi|} \to \Delta^{|\Pi| - 1}$ which derives a meta-game mixture strategy $\Pi^\sigma$ from the empirical payoff matrix $\cU$. \psro thus defines the following iterative procedure: at the $i$-th iteration, $\pi_i \gets \textsc{ABR}(\bar{\pi}, \Pi^{\sigma_{i-1}})$ is introduced to the policy population, with $\sigma_{i-1} \gets f(\cU)$ and $\bar{\pi}$ a randomly initialised policy. Starting from an arbitrary initial population $\Pi_0$ (typically a singleton $\{ \pi_0 \}$), \psro proceeds iteratively, until the {\sc ABR} operator fails to achieves a strictly positive payoff at that iteration. Finally, we note that while any MSS can be used, specific ones offer appealing properties. In particular, when {\sc NE} is used as the meta-strategy solution, \psro is known to converge to a NE of the {\em full} game. We refer to this implementation as \psronash for short.

\paragraph{Neural Population Learning}
\neupl \citep{liu2022neupl} differs from the iterative population learning procedure described thus far in two ways. First, the population of policies $\Pi$ is represented using a shared conditional network $\Pi_{\theta, \Sigma} = \{ \Pi_\theta(\cdot | o_{\le t}, \sigma_i); \sigma_i \in \Sigma \}$, with $\Sigma = \{ \sigma_i \in \Delta^{N-1} \}^{N-1}_{i=0}$ representing the adjacency matrix of an interaction graph \citep{marta} or equivalently, a set of meta-game mixture strategies. Second, the optimisation of the policy population proceeds concurrently, with policy $\Pi_\theta(\cdot | o_{\le t}, \sigma_i)$ maximising its expected returns against a mixture policy $\Pi^{\sigma_i}_{\theta, \Sigma}$, defined over the neural population itself. We note that $\Pi_\theta$  corresponds to a {\em single} neural network which is shared across all policies within the neural population. Extending MSS from the iterative case, \neupl updates the sequence of meta-strategies to best-respond to concurrently, using a meta-graph solver (MGS) $\cF$, with $\Sigma \gets \cF(\cU)$. Importantly, \neupl is known to converge to a NE when a NE meta-strategy solver is applied iteratively, with $\sigma_i \gets \textsc{Solve-NE}(\cU_{<i, <i})$. This replicates the strategic exploration dynamics of \psronash. While any MGS could be used in \neupl, we restrict our discussions to \psro (i.e. lower-triangular $\Sigma$s) for the rest of this work given its game-theoretic properties. Finally, we note that in contrast to the iterative setting, the {\em effective} population size is defined by the number of unique rows within $\Sigma$, with $|\textsc{UniqueRows}(\Sigma)| \le N$. The effective population size is therefore driven by the MGS used, adjusted dynamically through time based on the empirical payoff matrix $\cU$.

\section{Methods}
\label{sec:methods}

\begin{figure}
    \centering
    \includegraphics[width=0.9\columnwidth]{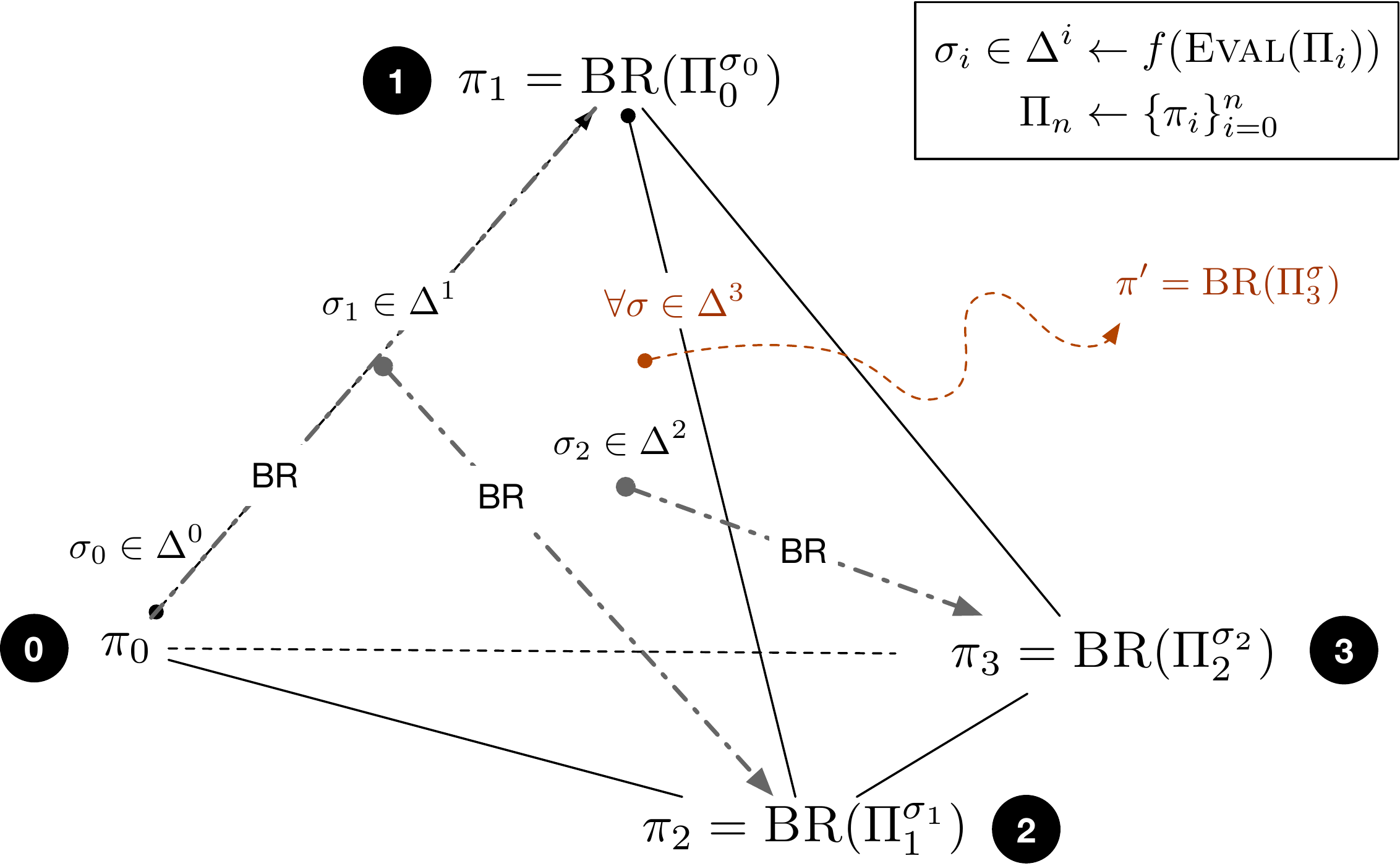}
    \caption{An iteratively expanding population where $\pi_i$ best-responds to $\Pi^{\sigma_{i-1}}_{i-1}$. Dashed arrows correspond to best-response operations. The solid lines are edges of the simplex and the vertices correspond to the set of basis policies. Each point $\sigma \in \Delta^3$ realises a mixture policy $\Pi^\sigma$ and admits its best-response policy $\textsc{BR}(\Pi^\sigma_3)$ (shown in brown).}
    \label{fig:population_simplex}
\end{figure}

\paragraph{Population Simplex} We now introduce the {\em population simplex} to serve as a geometric interpretation for population learning and motivates our proposed extension. A population simplex is defined by a set of $N$ polices $\Pi = \{\pi_i\}^{N-1}_{i=0}$ where each policy corresponds to a vertex of a simplex $\Delta^{N-1}$. Their mixture policies $\Pi^\sigma$ thus span the volume of the simplex with $\sigma$ a barycentric coordinate within the simplex. Analogous to \psro, the set of vertices are selected iteratively: i) given $\Pi$, a MSS selects a coordinate $\sigma \in \Delta^{|\Pi| - 1} \gets f(\textsc{Eval}(\Pi))$ and; ii) given $\Pi^\sigma$, a BR operator proposes a new vertex $\pi' \gets \textsc{BR}(\Pi^\sigma)$ that joins the existing population simplex to form an extended simplex $\Delta^{|\Pi|}$. At each iteration, \psro expands an $\Delta^{|\Pi|-1}$ to $\Delta^{|\Pi|}$ if $J(\pi', \Pi^\sigma) > 0$ with $\sigma \gets \textsc{Eval}(\Pi)$, if not, this iterative process terminates. This process is visualised in Figure~\ref{fig:population_simplex}, developing a sequence of policies iteratively starting from $\Pi = \{\pi_0\}$, forming a population 3-simplex. 

This geometric interpretation of population clarifies the interplay between the MSS and the BR operator --- for a given population simplex, one could in principle compute a best-response to each point within the simplex, developing infinitely many candidate policies that can be added to the population. Nevertheless, such procedure has been infeasible computationally, as best-response solving often comes at significant computational cost even for a single policy. A feasible solution is thus to rely on a meta-strategy solver. A MSS proposes a specific point within the simplex worth best-responding to, directing computational resources efficiently. This process forgoes optimal returns for all but a few select points for which the population offers best-responses, but yields population-level desiderata such as convergence to the NE \citep{mcmahan_planning_2003} or maximal exploration of the policy space \citep{balduzzi2019openended}.

\paragraph{Simplex Neural Population Learning}
Our proposal, is therefore to generalise \neupl to additionally and concurrently optimise best-responses to {\em all} mixtures within the population simplex. Specifically, we utilize \neupl to produce a set of basis policies and recognize that $\forall \sigma$ within the population simplex, we can optimise $\Pi_\theta(\cdot | o_{\le t}, \sigma)$ to maximise its expected returns against the mixture policy $\Pi^{\sigma}_{\theta, \Sigma}$. This leads to Algorithm~\ref{alg:simplex_neupl} where in addition to optimising the discrete set of conditional policies $\Pi_{\theta, \Sigma} \gets \{ \Pi_\theta(\cdot | o_{\le t}, \sigma_i) \}^{N}_{i=1}$ as in \neupl, we also optimise best-responses to any mixture policies $\Pi^\sigma$, with probability $\epsilon$, where the opponent prior $\sigma$ is sampled according to a symmetric Dirichlet distribution with equal concentration $\alpha$ assigned to each unique policies (i.e. $\textsc{UniqueRows}(\Sigma)$) of the neural population. In other words, we sample mixture opponent policies uniformly over the population simplex, with support over the set of unique policies in the population simplex. We denote the concentration parameters as $\bm{\alpha_\le}$ to indicate that $|\textsc{UniqueRows}(\Sigma)| \le N$ and $\textsc{Unif}(\cdot)$ refers to sampling one distribution from a set of probability distributions uniformly at random. This procedure is illustrated in Algorithm~\ref{alg:simplex_neupl}.

At convergence, simplex-\neupl leads to a conditional policy from which one may construct not only all mixture policies within the population simplex $\Pi^\sigma$, but also, their Bayes-optimal responses $\Pi_\theta(\cdot | o_{\le t}, \sigma)$. In subsequent analyses, we refer to $\Pi_\theta(\cdot | o_{\le t}, \bar{\sigma})$ as the {\em uninformed} policy with $\bar{\sigma}$ the uniform distribution (i.e. uninformative opponent prior) and $\Pi_\theta(\cdot | o_{\le t}, \sigma)$ the {\em informed} policy as it is conditioned on the (often privileged) opponent sampling distribution.

\begin{algorithm}[ht]
\caption{Simplex Neural Population Learning}
\label{alg:simplex_neupl}
\begin{algorithmic}[1]
    \State $\Pi_\theta(\cdot|o_{\le t},\sigma)$ \Comment{Conditional neural population net.}
    \State $\Sigma \eqdef \{\sigma_i\}^{N-1}_{i=0}$ \Comment{Initial interaction graph.}
    \State $\cF: \bR^{N \times N} \to \bR^{N \times N}$ \Comment{Meta-graph solver.}
    \While {continuing}
        \State $\np \gets \{\Pi_{\theta}(\cdot|s,\sigma_i)\}^{N-1}_{i=0}$ \Comment{Neural population.}
        \State $\textrm{simplex-sampling} \sim \textsc{Bern}(\epsilon)$ \Comment{With prob. $\epsilon.$}
        \If{$\textrm{simplex-sampling}$}
        \State $\sigma \sim \textsc{Dirichlet}(\bm{\alpha_\le})$ \Comment{From simplex.}
        \Else
            \State $\sigma \sim \textsc{Unif}(\textsc{UniqueRows}(\Sigma))$ \Comment{From MGS.}
        \EndIf
        \State $\Pi_\theta(\cdot|o_{\le t}, \sigma) \gets \textsc{ABR}(\Pi_\theta(\cdot| o_{\le t}, \sigma), \Pi^\sigma_{\theta, \Sigma})$
        \State $\cU \gets \textsc{Eval}(\np)$ \Comment{(Optional) if $\cF$ adaptive.}
        \State $\Sigma \gets \cF(\cU)$ \Comment{(Optional) if $\cF$ adaptive.}
    \EndWhile
    \State \Return $\Pi_\theta$, $\Sigma$
\end{algorithmic}
\end{algorithm}

\section{Results}
\label{sec:results}

We experiment with simplex-\neupl across two domains. First, we study the imperfect-information game of {\em goofspiel} which remains amenable to analytical posterior inference and exact best-response solving {\bf (Section~\ref{sec:goofspiel})}. Second, we explore the partially-observed, spatiotemporal strategy game of \rws, where information-seeking actions and observation history representation are critical in inferring opponent strategies and therefore, winning the game. The policy space of the latter is significantly larger and we seek to understand the trade-offs involved in the choice of policies at test time as well as the effect of unseen opponents on implicit posterior inference {\bf (Section~\ref{sec:rws})}. Throughout all experiments, we follow \psronash where an off-the-shelf \nash solver is used at each iteration of the meta-game solving, as in \citet{liu2022neupl}. The specific implementation of $\fpsron$ used as well as further details of our specific experimental setup are described in Appendix~\ref{app:experiments}.

\subsection{Goofspiel}
\label{sec:goofspiel}

\begin{figure*}
    \centering
    \includegraphics[width=\textwidth]{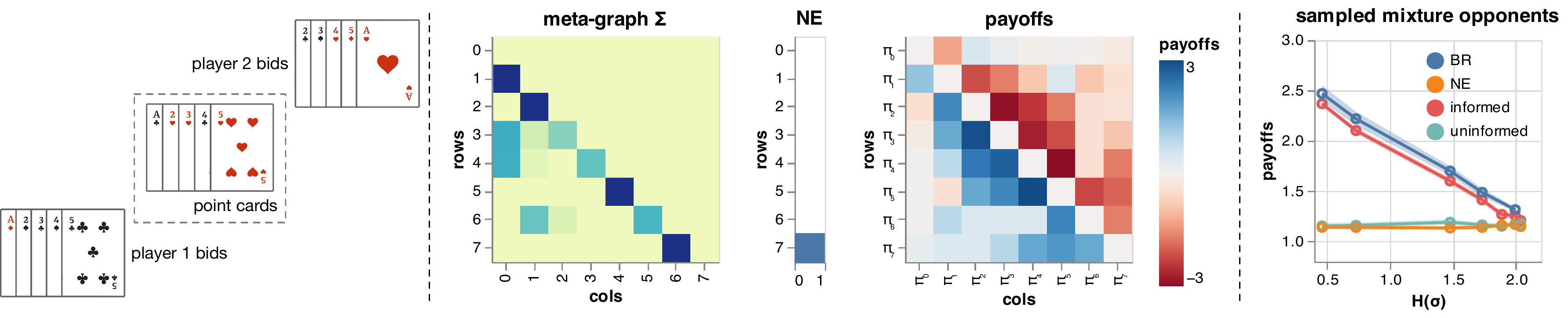}
    \caption{{\bf (Left)} an example game of {\em goofspiel} showing 5 point cards revealed in descending order and the two players each playing their bidding cards in (hidden) order. In this game player 1 wins the first point card but loses all subsequent points and ends up losing the game; {\bf (Middle)} a neural population of strategically diverse policies optimised by simplex-\neupl, following \psronash; {\bf (Right)} average return obtained by the exact best-response policy (blue), informed policy $\Pi_{\theta}(\cdot | o_{\le t}, \sigma)$ (red), uninformed policy $\Pi_{\theta}(\cdot | o_{\le t}, \bar{\sigma})$ (cyan) and the empirical NE mixture policy (orange) evaluated against 6 sets of opponent mixture policies $\{ \{ \Pi^{\sigma_{i, k}}_{\theta, \Sigma}; \sigma_{i, k} \sim \textsc{Dir}(\bm{\alpha}_k) \}^{256}_{i=1}; \bm{\alpha}_k \}^6_{k=1}$. The $k$-th opponent set features mixture policies whose mixture distributions are of a certain level of entropy (denoted $H(\sigma)$), sampled from a symmetric Dirichlet distribution of concentration $\bm{\alpha}_k$. Each point shows the average return against one of the opponent sets.}
    \label{fig:goofspiel_np_returns}
\end{figure*}

The game of {\it goofspiel} is a symmetric zero-sum bidding card game where players spend bid cards to collect points from a deck of point cards. In particular, we focus on the imperfect information variant of this game, with 5 point cards, revealed deterministically in descending order\footnote{Implementation from OpenSpiel, see Appendix~\ref{app:goofspiel_environment}}. Players do not observe the bidding card played by its opponent, but only the win-loss history of each point card. This game has long been subject to game theoretic analyses with well-known strategic cycles \citep{ross1971goofspiel, rhoads2012computer}. An example game is shown in Figure~\ref{fig:goofspiel_np_returns} {\bf (Left)} where player 2 wins the game by conceding the highest value point card but guarantees wins of all remaining point cards.

In this section, we empirically investigate the effect of simplex-\neupl in this domain. First, we show that simplex-\neupl preserves game-theoretic strategic exploration as in \citet{liu2022neupl}. This is expected, as any-mixture optimality implies that the resulting conditional policy $\Pi_{\theta}(\cdot | o_{\le t}, \sigma)$ can best-respond to the subset of mixture policies $\Sigma = \{\sigma_i\}^{N-1}_{i=0}$ recommended by the meta-graph solver. Second, we verify that more generally, the resulting {\em informed} policy approaches Bayes-optimality facing {\em any} mixture policies. This is a novel and important property for generalisation, as it allows for incorporating a wide range of prior beliefs at test time, as opposed to sampling from a small set of best-response policies enumerated during training. Lastly, we show that the resulting model exhibits implicit posterior inference over opponent identities through interaction. This echoes prior works showing that meta-learning over a range of tasks induces Bayes-optimal behaviours \citep{mikulik_meta-trained_2020, ortega_meta-learning_2019}. 

\paragraph{Game-Theoretic Strategic Exploration}
Strategic exploration in EGTA is typically expressed as iteratively solving for best-responses to mixture strategies of the induced empirical game, which are a subset of all mixture policies over the population of policies. 
Figure~\ref{fig:goofspiel_np_returns} {\bf (Middle)} illustrates an example neural population learned by simplex-\neupl, where each policy is optimised to best-respond to the NE over its predecessors as in \psronash. Specifically, the $i$-th policy $\Pi_{\theta}(\cdot | o_{\le t}, \sigma_i)$ is optimised to best-respond to a mixture over $\Pi^{\Sigma_{<i}}_{\theta} = \{\Pi_{\theta}(\cdot | o_{\le t}, \sigma_j)\}^{i-1}_{j=1}$, following the meta-strategy NE solver given their pairwise payoff matrix $\cU_{<i, <i}$. The initial policy of this policy population is fixed to play bid cards in random order with a known best-response that plays bid cards in descending order, spending bid cards matching the point card at each turn \citep{ross1971goofspiel}. Indeed, $\Pi_{\theta}(\cdot | o_{\le t}, \sigma_1)$ solely focuses on best-responding to the initial random policy and implements this deterministic point-matching policy. In turn, $\Pi_{\theta}(\cdot | o_{\le t}, \sigma_2)$ seeks to best-respond to $\Pi_{\theta}(\cdot | o_{\le t}, \sigma_1)$, sacrificing the highest value point card in exchange for all remaining points. This recovers the known optimal policy against the point-matching. We further illustrate the set of policies implemented by the neural population in Appendix~\ref{app:goofspiel_policies}. Extending \neupl, we confirm that simplex-\neupl similarly accommodates principled population learning algorithms such as \psronash, exploring the policy space of the game strategically.

\paragraph{Any-Mixture Bayes-Optimality}
We now verify empirically that our proposed extension leads to a conditional policy that can best-respond to {\em any} mixture policies supported by the neural population, by simply conditioning on the prior distribution over opponent identities. To this end, we sample arbitrary prior distributions $\sigma$ over the simplex from symmetric Dirichlet distributions and evaluate the expected returns achieved by our method and several baselines against the same mixture policies $\Pi_{\theta, \Sigma}^\sigma$. We compare {\bf i)} the {\em informed} policy $\Pi_{\theta}(\cdot | o_{\le t}, \sigma)$, conditioned on the true prior $\sigma$; {\bf ii)} the {\em uninformed} policy $\Pi_{\theta}(\cdot | o_{\le t}, \bar{\sigma})$, conditioned on an uninformative uniform prior $\bar{\sigma}$; {\bf iii)} an exact best-response policy solved analytically as well as {\bf iv)} the empirical NE mixture policy $\Pi_{\theta, \Sigma}^{\sigma_{\textsc{NE}}}$ with $\sigma_{\textsc{NE}} \gets \textsc{Solve-NE}(\cU)$.
Figure~\ref{fig:goofspiel_np_returns} {\bf (Right)} illustrates the result of this comparison, categorised by the levels of uncertainty present in the sampled priors. Notably, the conditional policy $\Pi_\theta(\cdot | o_{\le t}, \sigma)$ performs optimally against sampled mixture policies over the simplex, with its expected return approaching that of the exact best-response solution. Interestingly, we show that the uninformed policy performs markedly worse though the gap narrows as the true priors themselves become less informative with increased entropy. Last but not the least, we note that $\Pi_{\theta, \Sigma}^{\sigma_\textsc{NE}}$ achieved similar returns as $\Pi_\theta(\cdot | o_{\le t}, \bar{\sigma})$. This makes intuitive sense, as neither policy incorporates prior belief over the opponent distribution. Further details on the experimental setup, including the sampling of prior distributions and exact best-response solving are described in Appendix~\ref{app:goofspiel_any_mixture}.

\paragraph{Implicit Bayesian Inference of Opponent Strategies}

\begin{figure*}
    \centering
    \includegraphics[width=\textwidth]{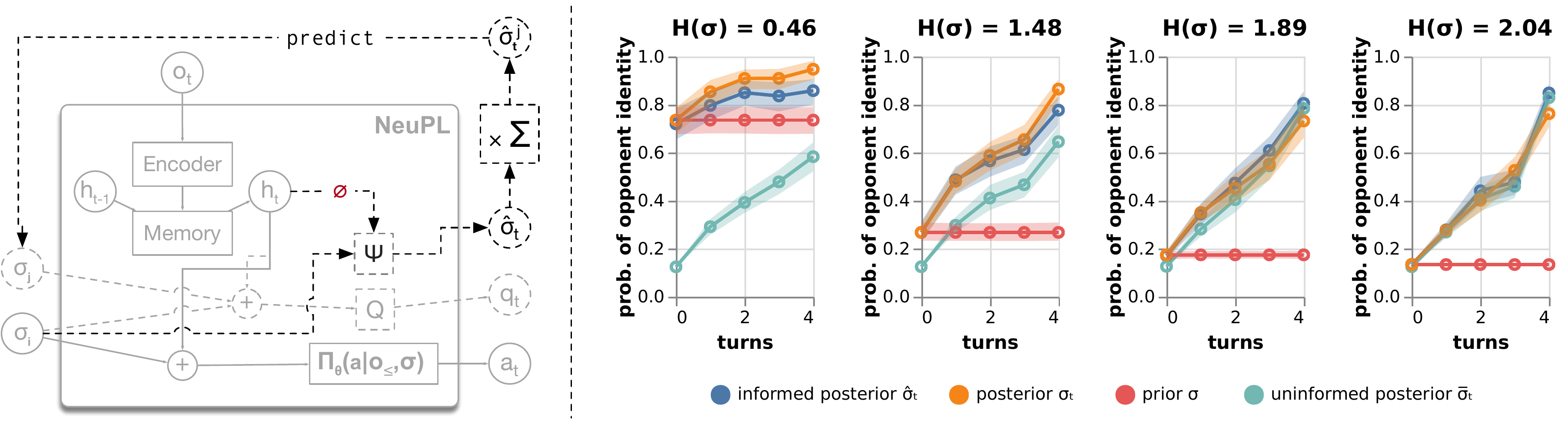}
    \caption{{\bf (Left)} The network architecture of simplex-\neupl with a posterior readout head. The baseline \neupl architecture is shown in gray as in \citet{liu2022neupl} and is used unmodified in this work; {\bf (Right)} The probability assigned to the true opponent identity under different distributions, including $\sigma$ the prior distribution, $\sigma_t$ the analytical posterior distribution, $\hat{\sigma}_t$ the ``implicit'' posterior distribution inferred by the posterior readout head and $\bar{\sigma}_t$ the ``implicit'' posterior distribution inferred with an uninformative uniform prior. The probability assignment is shown across ``turns'' (i.e. no card has been played at turn 0).}
    \label{fig:goofspiel_posterior}
\end{figure*}

For the instance of {\em goofspiel} that we are considering it is possible to compute the
posterior distributions over opponent identities analytically, given a prior distribution, a set of policies and an observation history. Consider a policy population $\Pi = \{ \pi_\theta(\cdot | o_{\le t}, \sigma_i)\}^{N-1}_{i=0}$ with $o_t = (a_t, w_t)$ where $a_t$ denotes the private bid card played by the player at time $t$, $w_t$ the publicly observed binary win-loss of the previous point card, $\sigma_i$ the identity of the player's policy, $\sigma_j$ the opponent identity at play, $\Pr(\sigma_j)$ the prior over the opponent identities, $a'_t$ the unobserved action taken by the opponent at time $t$. The posterior distribution over the opponent policy $\sigma_j$ can be computed by 
\small
$$\Pr(\sigma_j | o_{\le t}) = \frac{ \Pr(o_{\le t} | \sigma_j) \Pr(\sigma_j) }{\sum_{\hat{\sigma}_j} \Pr(o_{\le t} | \hat{\sigma}_j) \Pr(\hat{\sigma}_j)}$$
\normalsize with:
\small
$$
\Pr(o_{\le t} | \sigma_j) = \sum_{a'_{<t}} \Bigg[ \prod^{t-1}_{k=0} \pi_\theta(a'_k | o'_{<k}, \sigma_j) \Pr(w_{k+1}|a_k, a'_k) \Bigg]
$$
\normalsize
where the sum is over all legal unobserved opponent action sequences $a'_{<t}$. Specifically, the set of legal action sequences corresponds to all $5!$ permutations of the bidding cards. Note that as the underlying state transition function is deterministic, $\Pr(w_{k+1}|a_k, a'_k)$ reduces to the binary indicator on the consistency between the pair of bidding card at step $k$ and the publicly observed win-loss at step $k+1$. Intuitively, the analytical posterior distribution considers the likelihood of all possible unobserved opponent action sequences that are consistent with the public win-loss history.

To verify that the observation history representation implicitly performs inference of the opponent identity, we introduced a posterior readout network $\hat{\sigma}_t \gets \Psi(\sigma_i, \varnothing(h_t))$ that is optimised to infer the true opponent identity $\sigma_j$ given a prior distribution $\sigma_i$, as shown in Figure~\ref{fig:goofspiel_posterior} {\bf (Left)}. The readout head is parameterised by a MLP network and outputs a probability assignment $\hat{\sigma}_t$ over the set of meta-game strategies $\Sigma$. We report the output $\hat{\sigma}_t$ as the ``implicit'' posterior distribution at time $t$. Note that the readout head is optimised with a stop-gradient operator (shown as $\varnothing$) and has no influence on the representation of the policy network. The readout head is optimised alongside the agent during training, with an auxiliary regression loss.

Figure~\ref{fig:goofspiel_posterior} {\bf (Right)} illustrates the Bayesian posterior update that occurs implicitly through interaction within an episode. In particular, we report the probability assigned to the ground truth opponent identity under several distributions, grouped by different levels of entropy in the priors sampled from symmetric Dirichlet distributions. We make the following observations. First, both the informed implicit posterior and analytical posterior assign identical probability to the opponent identity as the prior distribution at turn 0, implying that the posterior readout correctly incorporates prior information about the opponent before any interaction. Similarly, the uninformed implicit posterior, $\bar{\sigma}_0 \eqdef \Psi(\bar{\sigma}, \varnothing(h_0))$ reliably assigns a probability consistent with that of a uniform distribution. As the entropy of the prior distribution increases and becomes less informative, all distributions converge to the same probability assignment at turn 0. Second, the implicit posterior distribution $\hat{\sigma}_t$ closely follows that of the analytical posterior $\sigma_t$ and tends to correctly assign higher probability to the true opponent policy as the policy gathers more evidence about its opponent. We note that the implicit posterior need not exactly reproduce its analytical counterpart, as an accurate prediction of the opponent identity is only needed if doing so improves the expected return of the policy. Lastly, we note that remarkably, an uninformed policy quickly catches up to its informed counterpart purely through interaction with the opponent, making increasingly accurate posterior inference about the opponent at play. We emphasise that representation of observation history, predictive of opponent identities, is solely a result of population learning. We visualise such implicit posterior inference in action in Appendix~\ref{app:goofspiel_posterior_in_action}.

\subsection{Running-with-Scissors}
\label{sec:rws}

\begin{figure*}
    \centering
    \includegraphics[width=\textwidth]{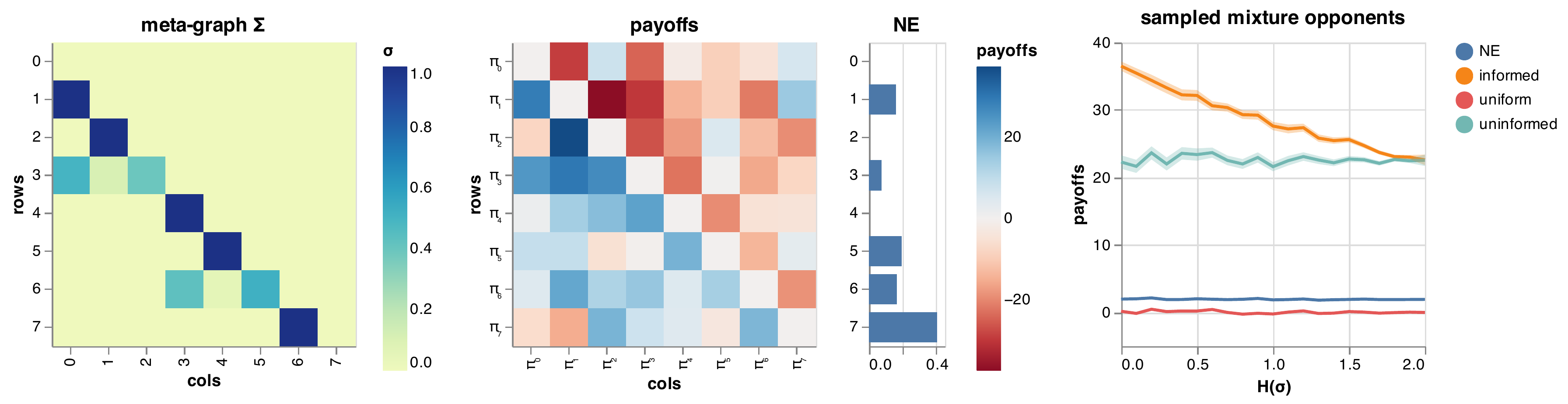}
    \caption{{\bf (Left)} the meta-graph $\Sigma$ representing the sequence of opponent mixtures to best-respond following NE; {\bf (Middle)} the pairwise payoff matrix between the population of 8 policies and their NE mixture $\sigma^{\textsc{NE}}$; {\bf (Right)} the expected payoffs of the NE mixture policy, uniform mixture policy, the informed policy $\Pi_\theta(\cdot | o_{\le t}, \sigma)$ and the uninformed policy $\Pi_\theta(\cdot | o_{\le t}, \bar{\sigma})$ evaluated against mixture opponent policies $\Pi^\sigma$ with $\sigma$ from Dirichlet distributions of different levels of concentration.}
    \label{fig:rws}
\end{figure*}

The game of \rws extends \rps to the spatiotemporal setting with two competing players collecting {\em rock}, {\em paper} and {\em scissors} such that one's inventory would compare favorably to its opponent's during the final confrontation. The game is partially-observed, with each player observing a small 4x4 grid in front of itself at each time. The two players confront each other when the episode terminates after 500 timesteps, or when one player {\em tags} its opponent during a close encounter. This game has been studied in-depth in prior works \citep{vezhnevets2020options, liu2022neupl}, revealing a range of interesting strategies and the importance of inferring opponent strategies through interaction in this game. In this section, we hope to understand the empirical {\em test-time} benefits of any-mixture optimality as enabled by simplex-\neupl. Further descriptions of the game is available in Appendix~\ref{app:rws_environment}.

We first study the scenario where both players have access to the same population of policies and illustrate that executing the NE mixture policy $\Pi^{\sigma_\textsc{NE}}_{\theta, \Sigma}$ is far from optimal when evaluated against arbitrarily sampled mixture policies. Instead, executing an uninformed policy that infer opponent strategy dynamically through interaction could be surprisingly effective. We then turn to the setting where the opponent, or ``column-player'', has access to a distinct, concealed population of policies. We investigate the mechanism of implicit posterior inference as implemented by the uninformed policy, showing that out-of-distribution opponent policies can be embedded in terms of strategically similar policies in one's own policy population.

\paragraph{Test-time Policy Selection}
Figure~\ref{fig:rws} {\bf (Left, Middle)} visualise a neural population of 8 policies, optimised via simplex-\neupl following \psronash --- the population represents a sequence of iterative best-responses, exhibiting several strategic cycles and does not admit a single dominant policy, as evidenced by its NE equilibrium $\sigma_\textsc{NE}$. Given this population of policies, Figure~\ref{fig:rws} {\bf (Right)} compares the expected returns of several polices that can be constructed using the same conditional network $\Pi_\theta(\cdot | o_{\le t}, \sigma)$, including $\Pi_{\theta, \Sigma}^{\sigma_\textsc{NE}}$ the NE mixture policy, $\Pi_{\theta, \Sigma}^{\bar{\sigma}}$ the uniform mixture policy, $\Pi_\theta(\cdot | o_{\le t}, \sigma)$ the policy informed of the (privileged) opponent mixture distribution and $\Pi_\theta(\cdot | o_{\le t}, \bar{\sigma})$, its uninformed counterpart. We make several observations. First, both uniform mixture policy and NE mixture policy significantly under-performed their adaptive counterparts, as they must sample and commit to a specific best-response policy without interaction that may or may not perform well against the specific opponent at play. Nevertheless, executing the NE mixture policy remains preferable, as it cannot lose to any mixture policies supported by the policy population and achieves a positive return in expectation against arbitrarily sampled mixture opponents $\Pi_{\theta, \Sigma}^\sigma$. 
The uniform-conditioned policy performs surprisingly well. In contrast to Figure~\ref{fig:goofspiel_np_returns}, the significant gap between $\Pi^{\sigma_{\textsc{NE}}}_{\theta, \Sigma}$ and the uninformed policy reflects the nature of this temporally-extended game: unlike {\em goofspiel} where each action has direct implication on the final return, \rws allows players to interact without committing to a specific strategy early in games. The ability to infer opponent identities through interaction is thus attractive in real-world games, many of which afford extended spatiotemporal structure.

\paragraph{Held-out Opponent Representation}

\begin{figure*}
    \centering
    \includegraphics[width=\textwidth]{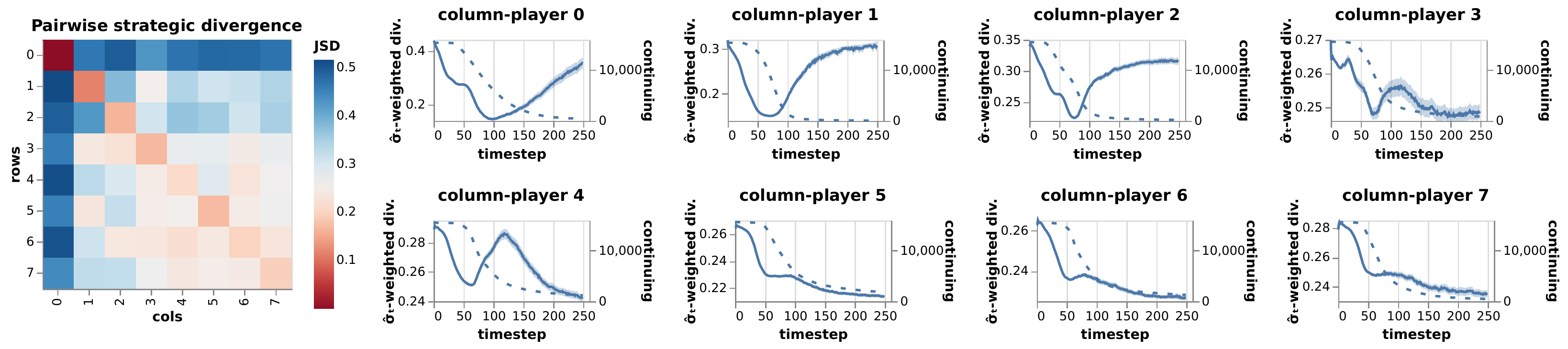}
    \caption{{\bf (Left)} Pairwise strategic divergence between row and column player's policies measured in Jensen-Shannon divergence; {\bf (Right)} Implicit posterior weighted strategic divergence inferred by an uninformed row-player policy when facing each column-player policy through time (solid) and the number of continuing episodes remaining due to early termination (dashed).}
    \label{fig:rws_jsd}
\end{figure*}

We have thus far restricted ourselves to the simplified setting where the ``column-player'' has access to the same population of policies as the ``row-player''. A more realistic setting, however, is one where the ``column-player'' has its own, concealed population of policies. In this setting, we study the behaviour of posterior inference as implemented by the uninformed row-player policy $\Pi_{\theta^r}(\cdot | o_{\le t}, \bar{\sigma})$, facing held-out column-player policies $\{ \Pi_{\theta^c}(\cdot | o_{\le t}, \sigma^c_i) \}^{N-1}_{i=0}$.

Figure~\ref{fig:rws_jsd} {\bf (Left)} visualise the pairwise strategic divergence matrix $\cD$ between the two populations of policies that share the same initial policy but are otherwise optimised independently, following \psronash. To measure behavioural similarity, we evaluate the expected Jensen-Shannon divergence between pairs of policies with element $\cD_{ij} = \expt_{o_{\le t} \sim \Tau^i} \Big[ D_{\textsc{JS}}[\Pi_{\theta^r}(\cdot | o_{\le t}, \sigma^r_i) || \Pi_{\theta^c}(\cdot | o_{\le t}, \sigma^c_j)] \Big]$ where $\Tau^i$ is the observation history distribution following $\Pi_{\theta^r}(\cdot | o_{\le t}, \sigma^r_i)$, playing against the uniform mixture policy of the column-player. We highlight the following observations on the two populations of policies. First, both populations developed the three pure-resource policies, echoing \citep{liu2022neupl}, as reflected by the comparatively reduced strategic divergence along the diagonal elements. This trend becomes less pronounced beyond the initial pure-resource policies, as the sequence of best-responses diverge across the two populations due to approximation in best-response solving. Figure~\ref{fig:rws_jsd} {\bf (Right)} illustrates the effect of inferred implicit posterior $\hat{\sigma}_t$ in the presence of unknown opponents: for the $j$-th column-player policy, the $\hat{\sigma}_t$-weighted divergence 
$\hat{\sigma}^T_t \cD_{[:,j]}$ decreases early on (shown in solid), as the row-player policy represents the column-player policies implicitly in terms of policies similar to that of its own policy population. As an increasing number of episodes early-terminate (due to players tagging each other, shown in dashed), the policy tend to struggle to identify the opposing strategies in the small number of continuing episodes.

\subsection{Ablation Studies}
\label{sec:ablation}

\begin{figure}
    \centering
    \includegraphics[width=\columnwidth]{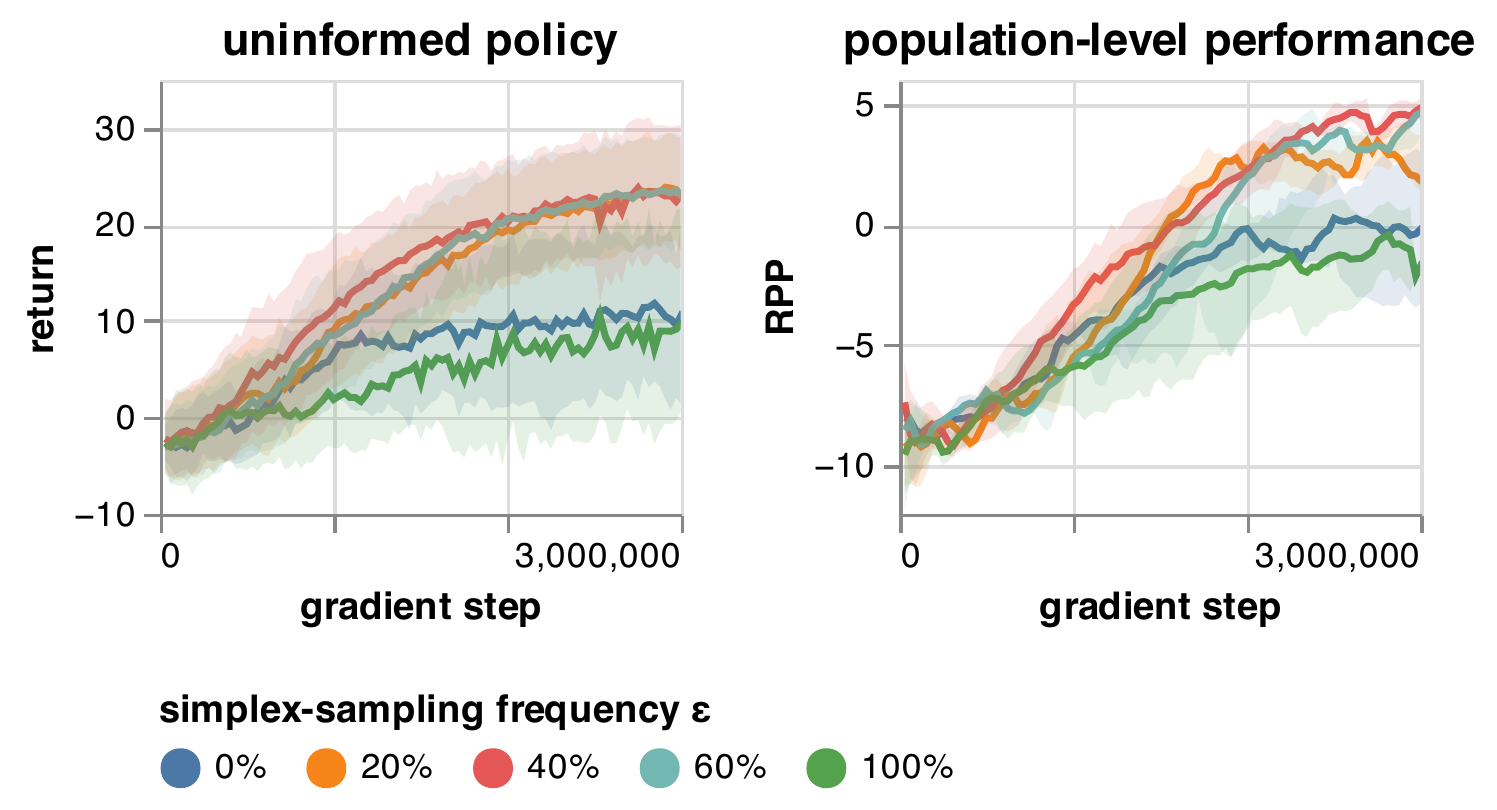}
    \caption{{\bf (Left)} expected returns of uninformed policies $\Pi_\theta(\cdot | o_{\le t}, \bar{\sigma})$ evaluated against a held-out uniform mixture of opponent policies; {\bf (Right)} relative population performance evaluated against the same held-out opponent population. Each configuration is repeated over 3 trails.}
    \label{fig:rws_ablation}
\end{figure}

We investigate the impact of simplex-sampling in Algorithm~\ref{alg:simplex_neupl}. In particular, we are interested in the two extremes: compared to \neupl, is simplex-sampling necessary for generalisation to any-mixture optimality; and conversely, is simplex-sampling, by itself, sufficient for strategic exploration. To answer these questions, we compare the performance achieved by the uninformed policy $\Pi_\theta(\cdot | o_{\le t}, \bar{\sigma})$ when different simplex-sampling frequency $\epsilon$ is used when training neural populations of up to 8 policies, evaluated against a uniform mixture of 8 held-out opponents in \rws. Figure~\ref{fig:rws_ablation} {\bf (Left)} reveals the importance of simplex-sampling in this setting. In particular, the uninformed policy fails to generalise optimally without sampling from the simplex during training. At the other extreme, sampling from the simplex {\em alone} significantly underperforms, too. We hypothesise that this is due to insufficient best-response learning, with few samples contributing towards learning strategically relevant policies following the MSS. 
Figure~\ref{fig:rws_ablation} {\bf (Right)} echoes this observation by evaluating populations of policies in relative terms (measured in Relative Population Performance, see Appendix~\ref{app:rpp}) against the same held-out opponent population. Interestingly, it shows that simplex-sampling can improve population-level performance by concurrently optimising a generalised class of best-response policies. This result corresponds to a generalised statement on knowledge transfer discussed in \citet{liu2022neupl} --- while \neupl enabled transfers across policies each best-responding to a specific mixture policy, simplex-\neupl encourages transfers across best-responses to {\em any} mixtures.

\section{Related Works}
\label{sec:related_works}

The geometric interpretation of mixed-strategies representation as points within a simplex dates as far back as 1951 when Nash Equilibria were shown to exist in finite games, using Brouwer's fixed-point theorem defined over any closed, bounded convex sets \citep{ne_jnash}. Our work proposes to consider a more dynamic view of policy population simplex --- one that is iteratively expanded with vertices corresponding to best-responses to specific points of the population simplex at the preceding iteration. Building on recent works on efficient representation of best-responses to mixed-strategies \citep{liu2022neupl}, our proposal renders it computationally feasible to learn best-responses to {\em any} points within the population simplex.

From the perspective of iterative game-solving, several prior works can be synergistic with simplex-\neupl, too. For instance, performing well against {\em any} mixtures over a set of policies has been previously studied in \citet{smith2020iterative, smith2020learning}, where approximate best-responses to mixture policies are constructed by combining Q-values of best-responses to individual mixture components. While the combined policy does not optimise a Bayes-optimal objective directly, such policies may prove beneficial when used as behaviour priors so as to accelerate best-response learning to any-mixture opponents. Motivated by similar observations as ours, \citet{wu2021l2e} proposed to study few-shot adaptation to diverse opponents via gradient-based meta-learning. By comparison, our proposed method yields a policy that adapts Bayes-optimally to a range of opponent priors, without further test-time gradient-based adaptation. While our work builds on prior works that suggest best-responding to mixed-strategies according to specific MSSs based on game-theoretic solution concepts \citep{lanctot2017unified, mcmahan_planning_2003}, other approaches have been recently proposed to use a learned MSS optimised for alternative population-level objectives \citep{feng2021neural}. \citet{McAleer2022Anytime} further proposed to consider an {\em expanded} restricted game such that the population as a whole is guaranteed to exhibit monotonically decreasing exploitability across best-response iterations. We leave further investigation in combining these ideas with simplex-\neupl, a general tool towards learning any-mixture best-responses, to future works.

On the other end of the spectrum, a rich body of prior work have also been dedicated towards understanding the role of representing uncertainties from the perspective of a single agent, similar to the role of opponent prior conditioning in our work. Such uncertainties may arise due to partial-observability of the underlying environment dynamics \citep{zintgraf2019varibad} or due to the presence of other agents in the environment \citep{vezhnevets2020options, raileanu2018modeling, zheng2018deep}. A common tool that cuts across these works is the explicit representation of latent variables designed to explain the variations in the observing player's observation history, including those caused by other interacting players. These works convincingly showed the need to reason about uncertainties in the environment explicitly, especially when interacting with other agents under partial-observability. Our learning method for each policy in the population is more closely aligned with the framework of Bayesian multi-task RL, where the policy learns to infer the underlying environment dynamics simply by optimising its expected returns on a distribution over tasks, without an explicitly designed latent variables representing such uncertainties. Consistent with prior empirical and theoretical studies \citep{humplik2019meta, ortega_meta-learning_2019, mikulik_meta-trained_2020}, we show that the learned observation history representation is predictive of the opponent identity, simply as a result of return maximization, without an auxiliary prediction task that encourages it to do so.

\section{Conclusions}
\label{sec:conclusions}

In this work, we interpret population learning geometrically and recognizes its connections to any-mixture Bayes-optimality. By learning best-responses to the entire population simplex, we obtain a conditional policy that can not only execute arbitrary policies within the simplex, but also, their Bayes-optimal responses. Empirically, we show that the resulting conditional policies are capable of incorporating a wide range of prior beliefs about the opponent, yielding near optimal returns against arbitrary mixture policies. Importantly, we show that for real-world games, an uninformed policy can be surprisingly effective, as it exploits the temporal structure of the game and optimally trades-off exploration and exploitation under uncertainty.

\section*{Acknowledgements}

We would like to thank Stephen McAleer, Ian Gemp and Karl Tuyls for helpful comments and suggestions on an early draft of this work. John Schultz and Edward Lockhart for their support and continued contributions to the OpenSpiel library which made various game-theoretic analysis possible in this work. Finally, we would like to thank the anonymous ICML reviewers for their thoughtful and constructive feedback on our work which help improved our submission significantly.

\bibliography{paper}
\bibliographystyle{icml2022}

\newpage
\clearpage
\appendix

\section{Results}

\subsection{Goofspiel}

\subsubsection{Environment Settings}

\label{app:goofspiel_environment}

The specific implementation of the game is available as part of OpenSpiel \citep{lanctot2019openspiel}, instantiated with the following game string: 

\small
\begin{verbatim}
 goofspiel(imp_info=true,
           egocentric=True,
           num_cards=5,
           points_order=descending,
           returns_type=point_difference))
\end{verbatim}
\normalsize

We consider the imperfection information, two-player zero-sum of {\it goofspiel} with 5 point cards revealed in descending order. At turn $t$, each player observes $o_t$ consisting of the revealed point card $p_t$, its action history $a_{<t} = (a_0, a_1, \dots, a_{t-1})$ as well as the win, loss and draw history $w_{<t}$ ($w_i \in \{-1, 0, 1\}$) of previously revealed point cards. The player then plays one of the remaining bidding cards $a_t$ from its hand. We denote all valid action history $\cA_{<t}$, corresponding to all permutations of $t$ cards from the initial hand of 5 bidding cards for each player.

\subsubsection{Strategic Exploration}
\label{app:goofspiel_policies}

\begin{figure}[ht]
    \centering
    \includegraphics[width=\columnwidth]{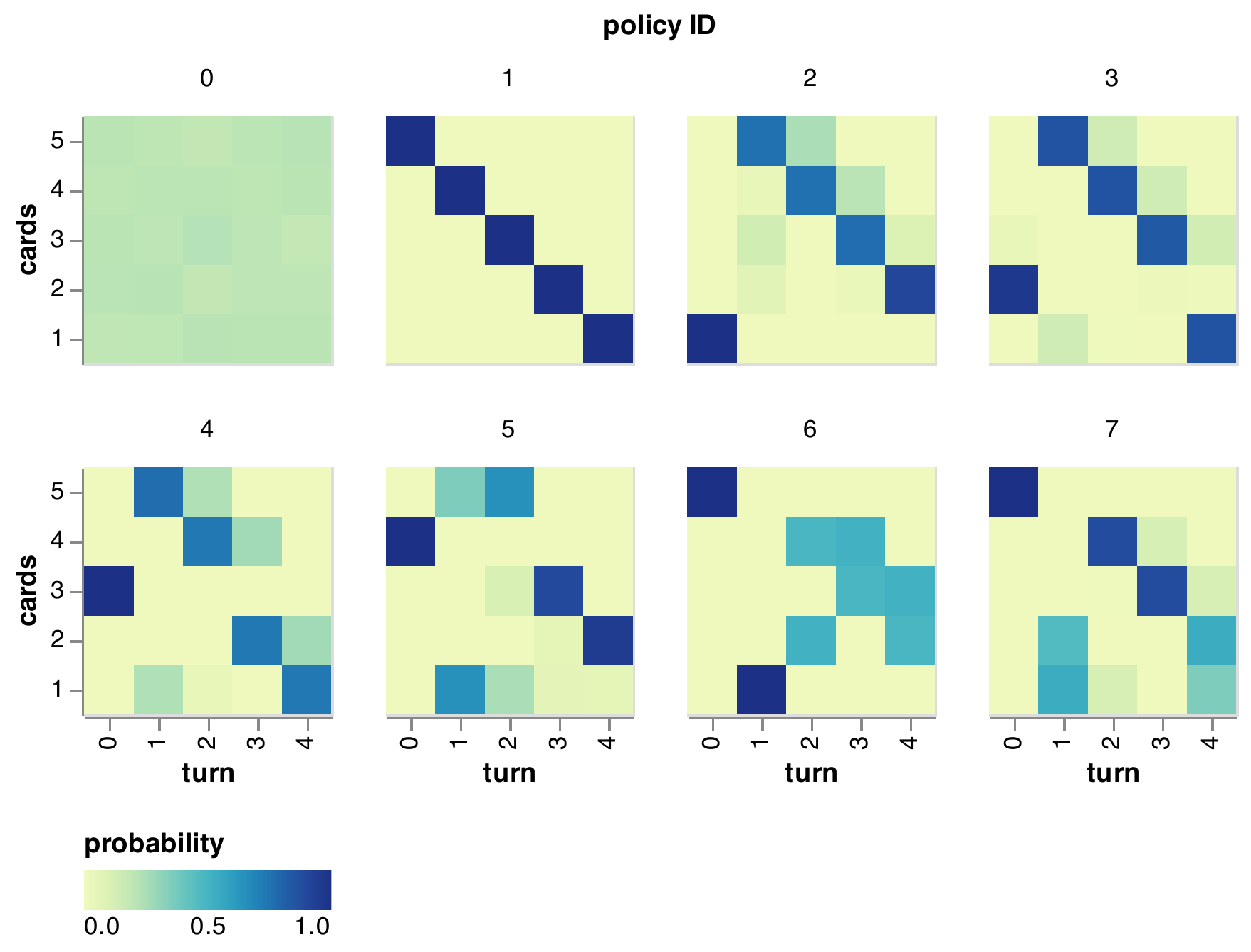}
    \caption{Policy profiles of a learned neural population $\{ \Pi_\theta(\cdot | o_{\le t}, \sigma_i), \sigma_i \in \Sigma \}$ in the game of {\em goofspiel}. The initial policy $\Pi_\theta(\cdot | o_{\le t}, \sigma_0)$ is fixed to play randomly. We simulate each pair of policies (with replacement) for 32 games and average the probability of playing each card at each turn, across all such pairs.}
    \label{fig:goofspiel_policy_profiles}
\end{figure}

We visualise the set of polices optimised via simplex-\neupl in the game of {\em goofspiel} described in Appendix~\ref{app:goofspiel_environment} in Figure~\ref{fig:goofspiel_policy_profiles}, following \psronash. We note that the policies are conditioned on observation-history and the action profiles visualised are averaged across all pairwise matches.

\subsubsection{Any-Mixture Optimality}
\label{app:goofspiel_any_mixture}

To establish the expected returns of different policies against arbitrarily opponent mixture policies, we instantiate $\Pi^\sigma$ with 256 $\sigma$ sampled from \textsc{Dirichlet} distributions across 7 levels of concentrations. This led to 7 sets of prior distributions with different levels of entropy $H(\sigma)$, ranging from $0.46$ to $2.03$. Note that a uniform distribution over a population of 8 policies corresponds to an entropy of $2.08$. Given the set of sampled mixture policies, we evaluate each candidate policy against each sampled mixture policy for 32 episodes, yielding expected returns as reported in Figure~\ref{fig:goofspiel_np_returns}.

For exact best-response solving in {\em goofspiel}, we resort to the open-source implementation of policy aggregator \url{open_spiel/python/algorithms/policy_aggregator.py} and \url{open_spiel/algorithms/best_response.h} from OpenSpiel \citep{lanctot2019openspiel} to compute a best-response policy against each of the sampled mixture policies.

\subsubsection{Implicit Posterior Inference in Action}
\label{app:goofspiel_posterior_in_action}

\begin{figure*}[ht]
    \centering
    \includegraphics[width=\textwidth]{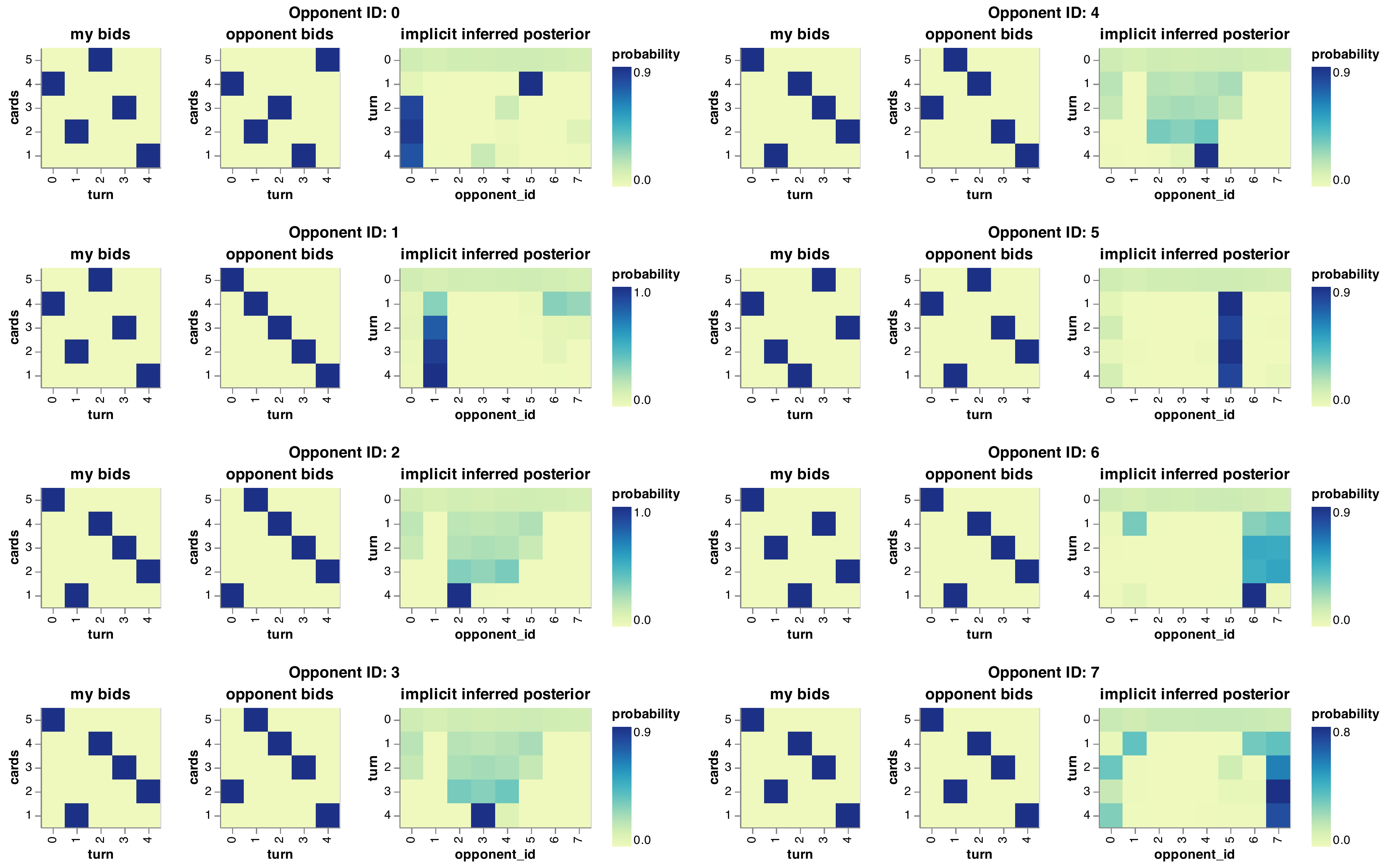}
    \caption{Example episodes where the uninformed policy $\Pi_\theta(\cdot | o_{\le t}, \bar{\sigma})$ plays against each one of the opponent policy. We visualise the bids from both players as well as the posterior inference from the first player's perspective. Note that players do not directly observe their opponents' previous bids but only observe the binary win-loss of previous point cards.}
    \label{fig:goofspiel_posterior_in_action}
\end{figure*}

Figure~\ref{fig:goofspiel_posterior_in_action} visualises 8 example episodes where the uninformed policy $\Pi_\theta(\cdot | o_{\le t}, \bar{\sigma})$ plays against each opponent in the policy population $\{ \Pi_\theta(\cdot | o_{\le t}, \sigma_i) \}^8_{i=0}$. We note that across all opponents, the implicit posterior as inferred from the internal observation-history representation of the agent is predictive of the true opponent identity, with increasing accuracy as the game progresses. We recall that players do not observe the actual bidding cards played by the opponent, but rather, the win-loss history of past point cards. This contributes to the uncertainty in opponent inference, in addition to the stochasticity present in the policy itself. Finally, we note that at turn 0, the implicit posterior readout reproduces the uniform prior, as expected.

\subsection{Running-with-Scissors}

\subsubsection{Environment}
\label{app:rws_environment}

\begin{figure}[ht]
  \begin{center}
   \includegraphics[width=\columnwidth]{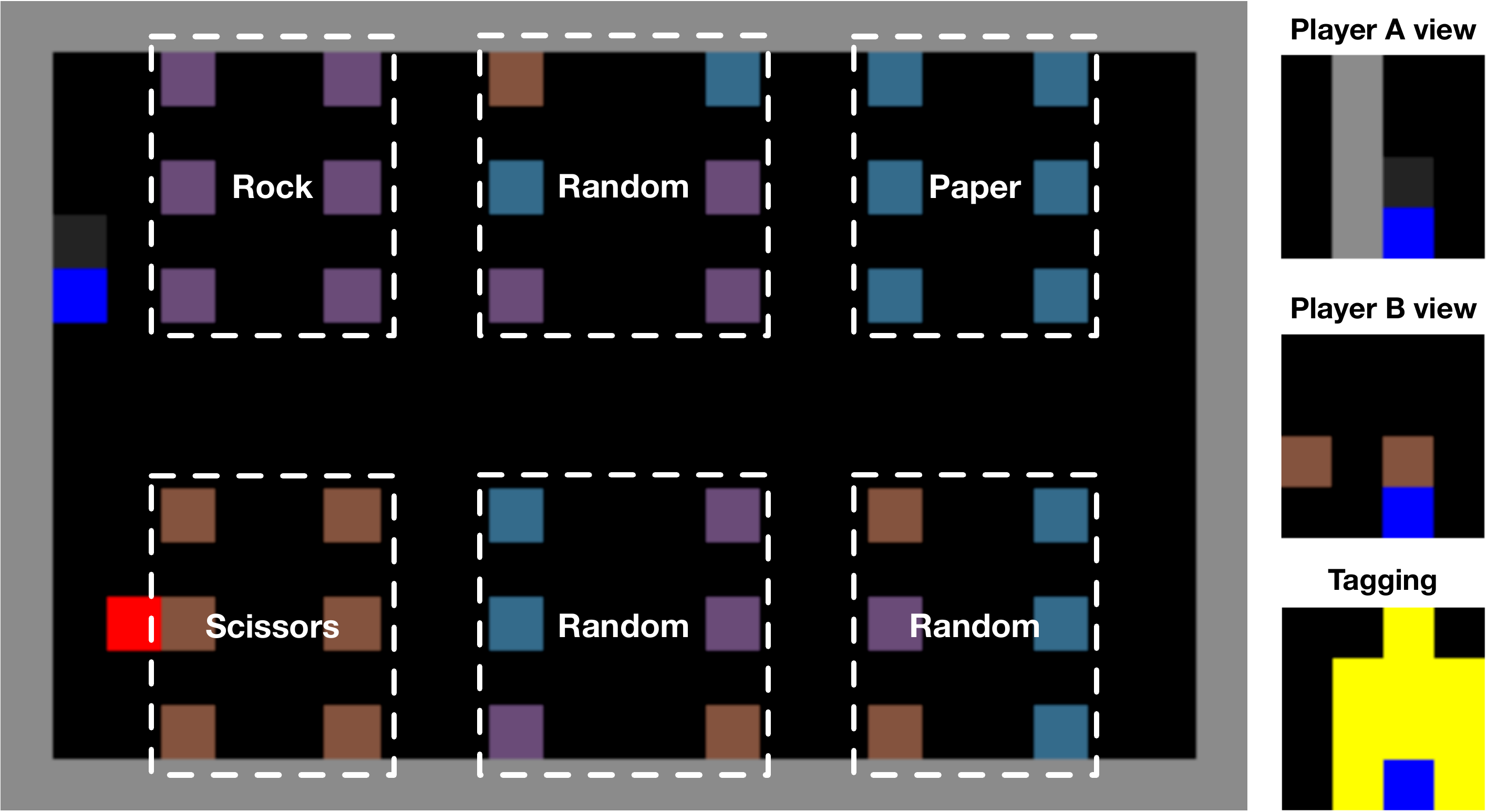}
  \end{center}
  \caption{An example frame of the \rws environment. Figure from \citet{liu2022neupl}.}
  \label{fig:rws_environment}
\end{figure}

Figure~\ref{fig:rws_environment} visualises an example scene of \rws during initialisation. The two players are randomly positioned and oriented in the game at the beginning of an episode with a 4x4 partial views of its surroundings. Each player maintains its own inventory, representing the proportion of resources in their possession, composed of {\em rock}, {\em paper} or {\em scissors} as visualised in coloured blocks. During a close encounter, each play may choose to tag its opponent, highlighting a small region in front of itself. If the opponent falls within the highlighted area, then the game resolves and the two players compare their inventories and receive rewards according to the \rps rules.

Similar to \citet{liu2022neupl}, we observed a range of strategic plays as explored by the neural population. This includes a policy that {\em observe-and-exploit} (shown as $\Pi_\theta(\cdot | o_{\le t}, \sigma_3)$ in Figure~\ref{fig:rws}) as well as a policy that plays deceptively to counter this policy.

\subsubsection{Relative Population Performance}
\label{app:rpp}

Relative Population Performance (RPP, \citet{balduzzi2019openended}) compares the performance of two populations in relative terms. Specifically, it computes the expected returns of one population versus another, if both parties play their respective NE mixture policies. We recall the formal definition of RPP.

\begin{definition}[Relative Population Performance]
Given two populations of policies $\fB, \fD$ and let $(\bf{p}, \bf{q})$ be a \nash equilibrium over the zero-sum game on $\cU_{\fB, \fD} \in \bR^{M \times N}$, {\rm Relative Population Performance} measures their relative performance: $v(\fB, \fD) \eqdef \bf{p}^T \cdot \cU_{\fB, \fD} \cdot \bf{q}$.
\end{definition}

\section{Experimental Setup}
\label{app:experiments}

\subsection{Agent Architecture}

The high-level agent network architecture is visualised in Figure~\ref{fig:goofspiel_posterior} which remains identical to that of \citet{liu2022neupl} besides the introduction of the posterior readout head. We recall that the readout head does not affect the representation learning of the main RL agent. 
Across both domains, we used the same MPO agent \citep{abdolmaleki2018maximum} as in \citet{liu2022neupl}, with 20 action samples drawn and evaluated by the learned Q-function per state at each gradient update. The target (Q-value and policy) networks are updated every 100 gradient steps. The policy head and Q-value networks are parameterised by MLPs of $(512, 256, 128, \textsc{NumActions})$ and $(512, 512, 128, 1)$ respectively with {\tt Elu} activation.

We describe domain-specific configuration below.

\subsubsection{{\em Goofspiel}}

In the game of {\em goofspiel}, we use a feed-forward agent without a memory component. This is possible because at each step, the environment observation contains the entire observation history for each player with perfect recall, rendering a recurrent network unnecessary. Instead of a recurrent network, we used a simple MLP network with 512 neurons for the {\tt Memory} module. The encoder consists of another MLP network of size $(128, 64)$, encoding the observation-history into a fixed size vector at each step. As is common in OpenSpiel \citep{lanctot2019openspiel}, the observation primarily consists of binary indicators of past events.

\subsubsection{\rws}

In \rws, the observation consists of a 4x4 pixel grid at each timestep as well as the player's current inventory. For observation-history encoding, we used a small convolutional network with a kernel shapes of $(1,)$ and 6 output channels followed by a MLP network of $(64, 64)$ dimensions. The pixel embedding is then concatenated with the inventory information and fed into another encoder parameterised by a 2-layer MLP network of size $(256, 256)$. This final per-timestep representation is then provided to a recurrent LSTM network, with a hidden size of 512. We note that the memory component is critically important in this game, as the observation at each timestep provides little information about the environment and the opponent. 

\subsection{Neural Population Configuration}

We used a maximum population size of 8 across all our experiments. We used an off the shelf linear program solver for {\textsc{Solve-NE}} and the iterative NE solving required by the meta-graph solver requires around 50ms to complete on a desktop CPU. In {\em goofspiel} we invoke the meta-graph solver every 10,000 gradient updates to allow time for the underlying RL agent to optimise. In \rws, we update the meta-graph every 1,000 gradient updates. We did not extensively adjust these hyper-parameters during our experimentation. 

\subsection{Meta-Graph Solver $\fpsron$}

\begin{algorithm}
  \caption{MGS implementing \psronash.}\label{alg:f_psro_nash}
  \begin{algorithmic}[1]
    \Function{$\fpsron$}{$\cU$}  \Comment{$\cU \in \bR^{N \times N}$.}
    \State Initialize $\Sigma \in \bR^{N \times N}$ with zeros.
    \For{$i \in \{1, \dots, N-1\}$}
    \State $\Sigma_{i+1,1:i} \gets \textsc{SOLVE-Nash}(\cU_{1:i, 1:i})$  
    \EndFor
    \State \Return{$\Sigma$}
    \EndFunction
  \end{algorithmic}
\end{algorithm}

As in \citet{liu2022neupl}, we iteratively apply \textsc{Solve-NE} \citep{shoham2008multiagent} to the sub-payoff matrices as the subsequent opponent mixture policy to best-respond to. This procedure is described in Algorithm~\ref{alg:f_psro_nash}.

\subsection{Training Setup}

Across both domains, we used a single TPU-v2 both to perform gradient updates for neural population of policies and to serve their inference requests during simulation. The game simulation is then performed on 256 remote CPU actors for \rws and 128 for {\em goofspiel}. Across both domains the neural population seem to have converged after 2M gradient updates (shared across all population members). This corresponds to about 1-days wall clock time for {\em goofspiel} and 3-days for \rws. During training, actors generate and write experience data to a replay server, with a maximum buffer size of 100,000 trajectories across both domains. Data are sampled uniformly from the replay server, without further prioritisation.

\end{document}